\documentclass[letterpaper]{article} 
\usepackage{aaai2027}  
\usepackage[hyphens]{url}  
\usepackage{graphicx} 
\usepackage{natbib}  
\usepackage{caption} 
\usepackage{algorithm}
\usepackage{algorithmic}

\usepackage{amsmath}

\usepackage{newfloat}
\usepackage{listings}
\DeclareCaptionStyle{ruled}{labelfont=normalfont,labelsep=colon,strut=off} 
\floatstyle{ruled}
\newfloat{listing}{tb}{lst}{}
\floatname{listing}{Listing}

\usepackage{booktabs}

\title{WorldMind: Decoupled Game World Model for State-Aware NPC Behavior}
 
\author{
    Zhiyang Deng\textsuperscript{\rm 1,\rm 2,$\dagger$},
    Boran Zhang\textsuperscript{\rm 1,$\dagger$},
    Danze Chen\textsuperscript{\rm 1,\rm 2,$\dagger$},
    Yeying Jin\textsuperscript{\rm 1,\rm 2,$\ddagger$}\corresponding
}
\affiliations{
    \textsuperscript{\rm 1}Tencent,
    \textsuperscript{\rm 2}National University of Singapore
}

\usepackage{amssymb} 
\usepackage{colortbl} 
\usepackage{multirow} 
\definecolor{oursrow}{RGB}{223,240,216}

\nocopyright

\usepackage{hyperref}
\makeatletter
\expandafter\let\csname ver@hyperref.sty\endcsname\relax
\makeatother
\definecolor{linkblue}{RGB}{20,86,180}
\hypersetup{
  hidelinks,
  breaklinks=true,
  pdftitle={WorldMind: Decoupled Game World Model for State-Aware NPC Behavior},
}

\begin{document}

\maketitle

\begingroup
\renewcommand{\thefootnote}{\fnsymbol{footnote}}

\footnotetext[2]{This work was completed during research internships at Tencent under the supervision of Yeying Jin.}
\footnotetext[3]{Project Lead.}

\endgroup

\begin{abstract}
Game world models have recently demonstrated promising capabilities in generating visually coherent and action-controllable gameplay videos. However, non-player character (NPC) behavior in existing models is either implicitly entangled with video generation or explicitly prescribed through external control signals. Consequently, a game world model has to jointly understand the state, plan the NPC's response and render its visual outcome, limiting its ability to produce responsive and state-aware NPC behavior. The challenge lies in the lack of an explicit interface for state-grounded decision-making. To this end, we introduce \textbf{WorldMind}, to our knowledge the first decoupled framework for state-aware NPC behavior in game world models. WorldMind separates interactive world modeling into four layers: an \emph{Understanding Layer} that constructs a compact state from generated frames; a \emph{Decision Layer} that reasons over the compact state to plan the NPC's next action; a \emph{Control Layer} that translates the actions into temporally aligned conditions; and a \emph{Generation Layer} that synthesizes their visual outcomes. By reconnecting layers in a closed interaction loop, WorldMind grounds NPC behavior in the evolving game state. We further introduce BOSS-140K, a dataset of gameplay videos paired with rich internal game states, together with an agent that automates the collection at scale. Experiments on BOSS-140K demonstrate reliable compact state reconstruction and mechanics-grounded planning, with WorldMind preferred over the baselines in approximately 70\% of pairwise comparisons for its more tactically appropriate and coherent NPC behavior.
Project page: \textcolor{linkblue}{\url{https://teawhite.cn/worldmind_projectpage/}}
\end{abstract}

\section{Introduction}

Recent game world models have demonstrated impressive visual fidelity and action controllability~\cite{gamengen,genie,matrixgame2}. Non-player characters (NPCs) are central to such worlds: they shape the dynamics and progression of gameplay by reacting to the current game state and the player's behavior. While player control has been studied extensively~\cite{diamond,oasis,incantation}, NPC behavior remains an underexplored aspect of interactive game world modeling.

\begin{figure*}[t]
    \centering
    \includegraphics[width=\linewidth]{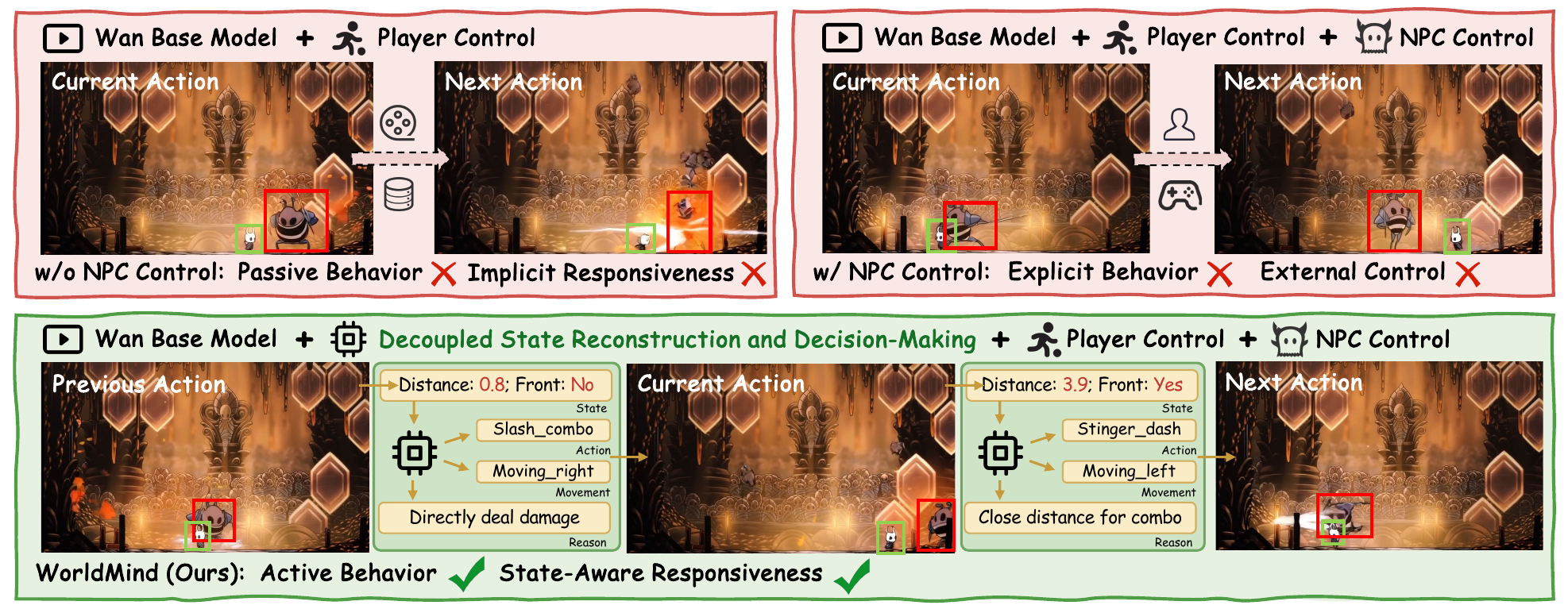}
    \vspace{-1.5em}
    \caption{Comparison with baseline NPC-control schemes.
    Using the same Wan base generator, the two baselines either leave the NPC
    action implicit in video generation (top left) or require it to be externally
    specified (top right). WorldMind instead constructs a compact state
    and reasons over it to plan the NPC's next action within the
    interaction loop, producing responsive, state-aware NPC behavior (bottom).
    \textcolor{green!50!black}{Green} and \textcolor{red!75!black}{red}
    bounding boxes denote the player and NPC, respectively.}
    \label{fig:teaser}
    \vspace{-1em}
\end{figure*}

As shown in Figure~\ref{fig:teaser}, existing game world models produce NPC behavior in one of two ways. In most, NPC behavior emerges implicitly from video generation, following behavioral patterns learned from the training data as part of the generated visual dynamics~\cite{combat}. The alternative exposes the NPC action as an external control signal and requires it to be specified outside the world model~\cite{incantation}. Neither formulation explicitly grounds the NPC's response in the evolving game state, including boss--player distance and relative angle, the boss's previous skill and per-skill cooldown availability. This limitation is especially consequential in boss fights, where planning an appropriate action requires accounting for a rapidly changing game state. Modeling responsive NPC behavior is therefore central to sustaining coherent and challenging gameplay.

The underlying difficulty lies in how NPC behavior is formulated: when NPC behavior remains implicit in video generation, a game world model must jointly infer the current state, reason about the NPC action and render its visual outcome. Because these functions differ in representation and temporal scope, combining them within a single generative process limits the responsiveness of NPC behavior. Addressing this entanglement requires an explicit interface that grounds NPC decisions in the evolving game state.
 
To this end, we introduce \textbf{WorldMind}, the first game world model to decouple state reconstruction and NPC decision-making from visual generation. Its framework structures the interaction loop so that the NPC action is determined from the compact state and translated into a generation condition before its visual consequence is synthesized. Accordingly, WorldMind comprises four layers: an \emph{Understanding Layer}, a \emph{Decision Layer}, a \emph{Control Layer} and a \emph{Generation Layer}. The \emph{Understanding Layer} captures what is happening during gameplay by reconstructing a compact state through complementary geometry and skill branches. The state is then passed to the \emph{Decision Layer}, which we instantiate with a general-purpose language model that reasons over it to plan the NPC's next action. The \emph{Control Layer} converts the actions into a temporally aligned conditioning sequence, supporting both direct control and high-level director control. The \emph{Generation Layer} leverages a video diffusion model to generate the resulting gameplay in real time. Generated frames return to the \emph{Understanding Layer}, closing the loop.


\begin{figure}[t]
    \centering
    \includegraphics[width=\linewidth]{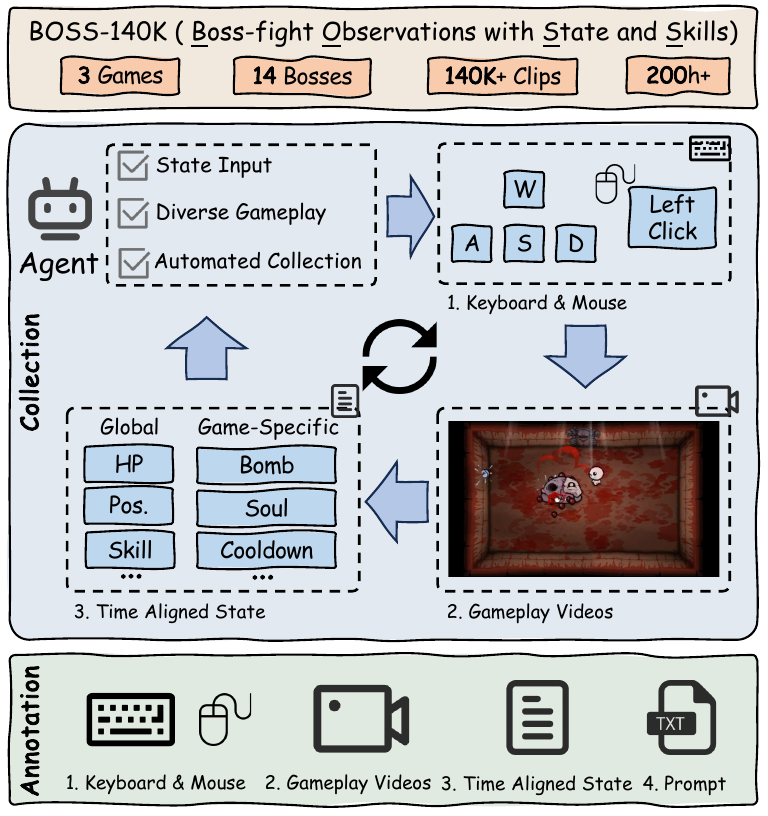}
    \vspace{-1em}
    \caption{Automated construction of BOSS-140K.
    A state-conditioned gameplay agent combines engine-internal state with
    game-specific strategies to collect diverse interactions without manual
    operation. The pipeline records gameplay video, internal state, and keyboard/mouse input on a shared timeline to produce frame-aligned annotations (see above).}
    \label{fig:dataset}
    \vspace{-1em}
\end{figure}

Training and evaluating such a system requires supervision beyond gameplay pixels and player inputs~\cite{minerl,vpt,egocs}. We therefore introduce \textbf{BOSS-140K} (\emph{\underline{B}oss-fight \underline{O}bservations with \underline{S}tates and \underline{S}kills}), a boss-fight dataset that pairs gameplay video with frame-aligned player controls, NPC skills and rich internal state annotations. To collect diverse interactions at scale, we develop a state-conditioned gameplay agent and a fully automated pipeline. Details can be seen in Figure~\ref{fig:dataset}.


In summary, our contributions are as follows:

\begin{itemize}
\item We introduce \textbf{WorldMind}, to our knowledge the first game world model to enable NPCs to make state-aware decisions as the game state evolves, rather than having their behavior emerge from video generation or be externally specified. WorldMind supports closed-loop, real-time interactive gameplay at approximately 20 FPS.

\item We propose a \textbf{four-layer decoupled framework} that separates state understanding, NPC decision-making, action conditioning, and visual synthesis. This design makes the NPC action an explicit decision grounded on the compact state, while supporting both direct player control and high-level director control within a unified framework for interactive gameplay generation.

\item We introduce \textbf{BOSS-140K}, a boss-fight dataset that pairs over 200h gameplay video with rich internal game-state annotations and captions. To construct BOSS-140K at scale, we further develop a gameplay agent conditioned on the internal game state and an automated data-collection pipeline that records diverse player-boss interactions without manual operation.

\end{itemize}

\section{Related Work}

\noindent \textbf{Interactive World Models.}
Recent diffusion and transformer architectures have driven rapid progress in video generation~\cite{svd,dit,cogvideox,hunyuanvideo}. In parallel, world models learn action-conditioned dynamics for prediction and control~\cite{worldmodels,dreamer,dreamerv2}. Game world models combine these capabilities to generate controllable gameplay rollouts from action inputs~\cite{gamengen,diamond,genie,oasis,gamefactory,scope}. Recent systems extend these rollouts to longer horizons and real-time streaming interaction~\cite{gamengen,matrixgame2}. To move beyond traditional control interfaces, recent work adopts natural language as an expressive action interface for compositional multi-entity control~\cite{incantation}.

\noindent \textbf{NPC Behavior Modeling.}
Conventional NPC behavior is authored with finite-state machines and behavior trees that map engine state to scripted actions~\cite{bt_survey,bt_intro}. LLM-based agents support more open-ended reasoning and planning~\cite{generativeagents,voyager,llmgame_survey}, but both paradigms assume structured observations and executable action interfaces. Video-based world models instead absorb NPC behavior into predicted visual dynamics or accept externally provided actions. COMBAT learns NPC behavior implicitly from single-player data~\cite{combat}, whereas ReactiveGWM conditions generation on high-level tactical labels~\cite{reactivegwm}; action selection therefore remains coupled with generation or specified externally. WorldMind constructs a compact state from generated observations and action history, explicitly plans the NPC's next action and separately renders its outcome.

\noindent \textbf{Datasets for Game World Models.}
Existing datasets for learning game dynamics commonly pair gameplay video with player controls, demonstrations or action annotations~\cite{ale,minerl,vpt,minedojo,nitrogen,egocs,openp2p}, supporting controllable generation but offering limited supervision for state-dependent NPC decisions. WildWorld adds explicit internal state~\cite{wildworld} but is not organized around NPC action decision in boss encounters. BOSS-140K fills this gap by pairing gameplay video with frame-aligned player controls, boss skills, animation states, rich engine variables and natural-language captions. Its state-conditioned collection agent adapts its strategy to elicit diverse boss responses.

\section{Method}

\begin{figure*}[t]
    \centering
    \includegraphics[width=0.97\linewidth]{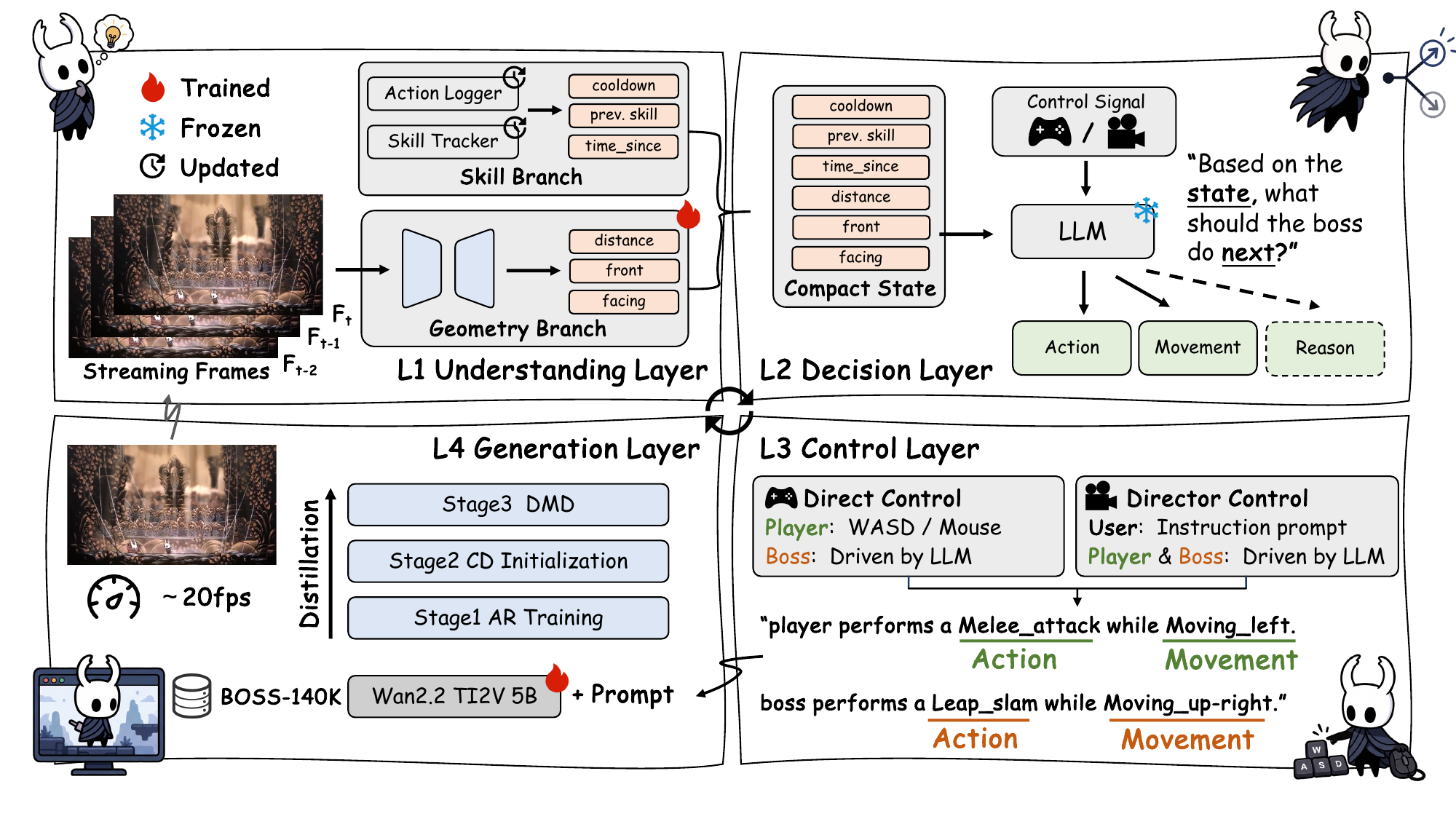}
    \vspace{-1em}
    \caption{Overview of the WorldMind framework.
    WorldMind connects four decoupled layers in a closed interaction loop. From
    the generated frames and boss action history, L1 constructs a compact state
    and passes it to L2, which reasons over the compact state and skill mechanics
    to plan the boss's next actions.
    L3 pairs the resulting boss action with player behavior obtained through
    either Direct or Director Control and converts both into a temporally aligned
    compositional natural-language conditioning sequence. Guided by this sequence,
    L4 synthesizes the next gameplay segment, whose frames are fed back to L1 to
    begin the next cycle of understanding, reasoning, and generation.}
    \label{fig:method}
    \vspace{-1em}
\end{figure*}

\subsection{Decoupled GWM Architecture}

Existing game world models typically model NPC behavior either implicitly through video generation or explicitly through an external control signal, leaving no explicit interface for deciding how the NPC should respond to the evolving game state. Establishing this interface raises three questions: \emph{what is happening}, \emph{what should the NPC do}, and \emph{how can state understanding, NPC decision-making, and video generation work together in a coherent closed-loop system}?

Guided by these questions, WorldMind decomposes interactive world modeling into four layers, as illustrated in Figure~\ref{fig:method}. The L1 Understanding Layer constructs a compact state by combining geometry variables predicted from generated frames with skill-history variables derived from the boss action history. The L2 Decision Layer reasons over this compact state to plan the NPC's next action. The L3 Control Layer translates the NPC and player actions into generation conditions used by the L4 Generation Layer to synthesize the corresponding visual outcomes. The generated frames are then fed back to L1 Understanding Layer, completing the interaction loop. This design exposes both the compact state and the NPC action while retaining a unified closed-loop system. We detail each layer below.

\subsection{L1: Understanding Layer}

Grounding an NPC decision in the evolving interaction requires recovering decision-relevant information about the current game state. This raises three questions: \emph{which variables constitute the game state}, \emph{what observable representations reveal them} and \emph{which variables matter for decision-making}? L1 addresses them through complementary skill and geometry branches, then selects the decision-relevant variables that form the compact state for L2 Decision Layer.

\noindent\textbf{Game state.}
We use game state to broadly denote the variables that characterize the evolving interaction, including entity attributes and status, action histories, skill availability, spatial relations, encounter phase and environmental context.

\noindent\textbf{Skill branch.}
The skill branch derives skill-history variables from NPC's action history. Its \emph{Action Logger} records each issued boss skill and timestamp, while the \emph{Skill Tracker} derives the previous skill, per-skill elapsed time and cooldown availability. The branch is deterministic and thus requires no engine state at inference time.

\noindent\textbf{Geometry branch.} 
Reasoning about the next action also depends on the boss--player spatial configuration: the same skill can have different tactical value depending on distance, relative angle and facing. Since these quantities cannot be recovered from the boss action history, the geometry branch infers them from a strictly causal window of generated RGB frames. A shared convolutional encoder extracts per-frame features, which a single-layer GRU aggregates over time. Conditioned on a learned boss-identity embedding, prediction heads produce continuous and discretized distance and angle, facing and other tactical attributes. The branch is trained with engine-logged geometry annotations but uses only generated frames and the boss identity at inference time.

\noindent\textbf{Compact state.}
The two branches yield candidate state variables whose relevance to downstream decisions varies. Based on a variable-selection analysis, L1 retains the most informative skill-history and geometry variables and adds the boss identity $\mathrm{id}^{b}$:
\begin{equation}
    \mathbf{s}_{t}
    =
    \big[
        \mathrm{id}^{b},
        \widetilde{\mathbf{c}}_{t},
        \widetilde{\mathbf{g}}_{t}
    \big],
\end{equation}
where $\widetilde{\mathbf{c}}_{t}$ and $\widetilde{\mathbf{g}}_{t}$ denote the selected skill-history and geometry variables. The resulting decision-relevant compact state $\mathbf{s}_{t}$ is then passed to L2 Decision Layer.

\subsection{L2: Decision Layer}

The compact state describes what is happening, but it does not by itself specify how the boss should respond. Making an appropriate decision and forming a tactically coherent plan require reasoning about how the mechanics of each skill relate to that state. Because these mechanics are naturally expressed in text, we instantiate L2 with a general language model that jointly reasons over the compact state and skill mechanics to form a short-horizon action plan. We deliberately DO NOT fine-tune the language model on BOSS-140K, avoiding direct imitation of the dataset's action distribution.

At each decision step $t$, L2 presents the compact state $\mathbf{s}_{t}$ and the boss's skill set $\mathcal{S}_{t}$ to the language model in textual form. Each skill is accompanied by a natural-language description of its mechanics and by cooldown information from L1's \emph{Skill Tracker}, including its cooldown and the time elapsed since it was last used. This allows the model to reason about when each skill can fire without pre-filtering the skill set. Under Direct Control, the model forms a boss plan $\mathbf{p}^{b}_{t}$. Under Director Control, it additionally conditions on a high-level instruction $\mathbf{d}$ and returns a player plan $\mathbf{p}^{p}_{t}$ alongside the boss plan. In either mode, it may also produce a rationale $r_t$:
\begin{align}
    \left(\mathbf{p}^{b}_{t},r_t\right)
    &=\mathrm{LLM}\!\left(\mathbf{s}_{t},\mathcal{S}_{t}\right)
    \quad\makebox[8.5em][r]{(Direct Control)}, \notag\\
    \left(\mathbf{p}^{p}_{t},\mathbf{p}^{b}_{t},r_t\right)
    &=\mathrm{LLM}\!\left(\mathbf{s}_{t},\mathcal{S}_{t},\mathbf{d}\right)
    \quad\makebox[8.5em][r]{(Director Control)}, \notag\\
    \mathbf{p}^{b}_{t}
    &=
    \left(
        \mathbf{a}^{b,(1)}_{t},
        \ldots,
        \mathbf{a}^{b,(H_t)}_{t}
    \right).
\end{align}
Each complete boss action $\mathbf{a}^{b,(j)}_{t}$ contains a skill $\ell^{(j)}_{t}$ and a movement directive. Its execution duration is attached deterministically as $\delta^{(j)}_{t}=D_{\mathrm{dur}}(\ell^{(j)}_{t})$ rather than predicted by the language model.
A deterministic check then simulates the proposed boss plan against a copy of the \emph{Skill Tracker}'s cooldown and removes any action that cannot legally fire.

We follow a receding-horizon strategy: only the first action of each plan is passed to L3, while the remaining actions are treated as provisional. The optional rationale $r_t$ is retained only for qualitative analysis and not passed onward.

\subsection{L3: Control Layer}

The output of L2 is defined at the level of complete entity actions, specifying what skill to execute, how to move and how long the skill lasts. L4, by contrast, expects a control prompt at every fixed temporal generation slot. L3 resolves this mismatch by converting complete actions into slot-aligned compositional prompts. It also supports two ways of specifying player behavior through Direct and Director Control.

\noindent\textbf{Natural-language action interface.}
Following prior work~\cite{incantation}, we use natural language as a structured action interface. It assigns a single action phrase to each entity at every temporal slot. In games, however, skill execution and movement are distinct control dimensions that often occur concurrently. Collapsing them into an undifferentiated action phrase can conflate the two signals. We therefore represent each entity with an \emph{``[ACTION] while [MOVEMENT]''} template, preserving both dimensions while retaining a purely textual interface and finer controllability.

\noindent\textbf{Temporal alignment.}
Action planning and video generation operate at different temporal scales. For each issued action, L3 converts its fixed execution duration into the corresponding number of L4 temporal slots and repeats the instantiated action--movement phrase across that interval. At every slot, the player and boss phrases are paired into a single compositional prompt. The resulting sequence contains one prompt per latent frame and provides the temporally aligned conditioning prompt sequence consumed by L4.

\noindent\textbf{Direct Control.}
In Direct Control, the user controls the player through keyboard and mouse inputs. At each temporal slot $k$, L3 maps the keyboard and mouse input events to the action and movement components of a structured player phrase. It then pairs this phrase with the duration-aligned boss phrase derived from the first action of $\mathbf{p}^{b}_{t}$. The resulting prompts preserve low-level player control while allowing the boss to act responsively.

\noindent\textbf{Director Control.}
In Director Control, the user replaces low-level inputs with the high-level natural-language instruction $\mathbf{d}$, such as \emph{"the player keeps evading while the boss remains highly aggressive"}. Conditioned on $\mathbf{d}$, The Decision Layer produces the player and boss plans $\mathbf{p}^{p}_{t}$ and $\mathbf{p}^{b}_{t}$, allowing the instruction to specify the desired interaction rather than exact controls. The Control Layer applies the same temporal projection and compositional interface to the first action of each plan before passing the resulting sequence to L4.

\subsection{L4: Generation Layer}

The conditioning sequence from L3 specifies what the player and boss should do, but not how these actions unfold visually. L4 generates the gameplay video conditioned on these prompt sequences, whose frames are fed back to L1 as new observations. It must therefore preserve scene continuity and faithfully reflect both entities' actions while operating fast enough to sustain continuous interaction.

\noindent\textbf{Base Model.}
We instantiate L4 with Wan~2.2 TI2V-5B~\cite{wan} and fine-tune it end-to-end on BOSS-140K with a batch size of 16 and 70k training steps. During data preprocessing, the frame-aligned player and boss actions are converted into temporally aligned conditioning sequences using the same format as L3. This adapts the pretrained video prior to boss-fight dynamics and our action interface.

\noindent\textbf{Distillation.}
The bidirectional model remains too slow for real-time interaction. We follow the three-stage Causal Forcing pipeline~\cite{causalforcing,causalforcingpp} to distill it into a few-step causal autoregressive generator through causal AR-diffusion teacher training, causal consistency distillation for initialization, and asymmetric DMD~\cite{selfforcing,dmd,dmd2}. The distilled generator runs at approximately $20$~FPS, supporting real-time gameplay.

\subsection{Closed-Loop Interaction}

The four layers form an iterative interaction cycle rather than a one-way generation pipeline. After L4 produces a segment, its frames become the next visual observations for L1, while the executed boss action updates the skill history. L1 then constructs a new compact state and L2 replans from the updated interaction context. Because only the first planned action for each model-controlled entity is committed, subsequent decisions can respond to the visual consequences of earlier ones instead of following a fixed open-loop sequence. Thus, the layers remain decoupled in function but are coupled through feedback over time.

\subsection{BOSS-140K Dataset}

WorldMind exposes the compact state and boss actions as explicit variables, requiring supervision beyond standard gameplay recordings. We introduce \textbf{BOSS-140K} (\emph{\underline{B}oss-fight \underline{O}bservations with \underline{S}tates and \underline{S}kills}), a game world model dataset that provides this supervision by pairing gameplay video with frame-aligned player controls, boss actions, engine-internal state variables and natural-language captions. It contains $144{,}631$ clips totaling over $200$ hours from 14 bosses across \emph{Game A}, \emph{Hollow Knight} and \emph{The Binding of Isaac} (hereafter \emph{Isaac}), spanning 2.5D and 2D perspectives. We refer to one title as \emph{Game A} and blur all of its frames.

Collecting BOSS-140K poses a coverage challenge as well as a scale challenge: unguided play under-samples the state-dependent interactions that WorldMind must model. We therefore build a fully automated pipeline around a state-conditioned gameplay agent, as illustrated in Figure~\ref{fig:dataset}. The agent uses engine-internal state and game-specific strategies to adapt its behavior to the evolving game state, while the pipeline records all supervision signals on a shared timeline. Engine state is used only for data collection and is unavailable to WorldMind at inference time. 

\begin{figure*}[t]
    \centering
    \includegraphics[width=\linewidth]{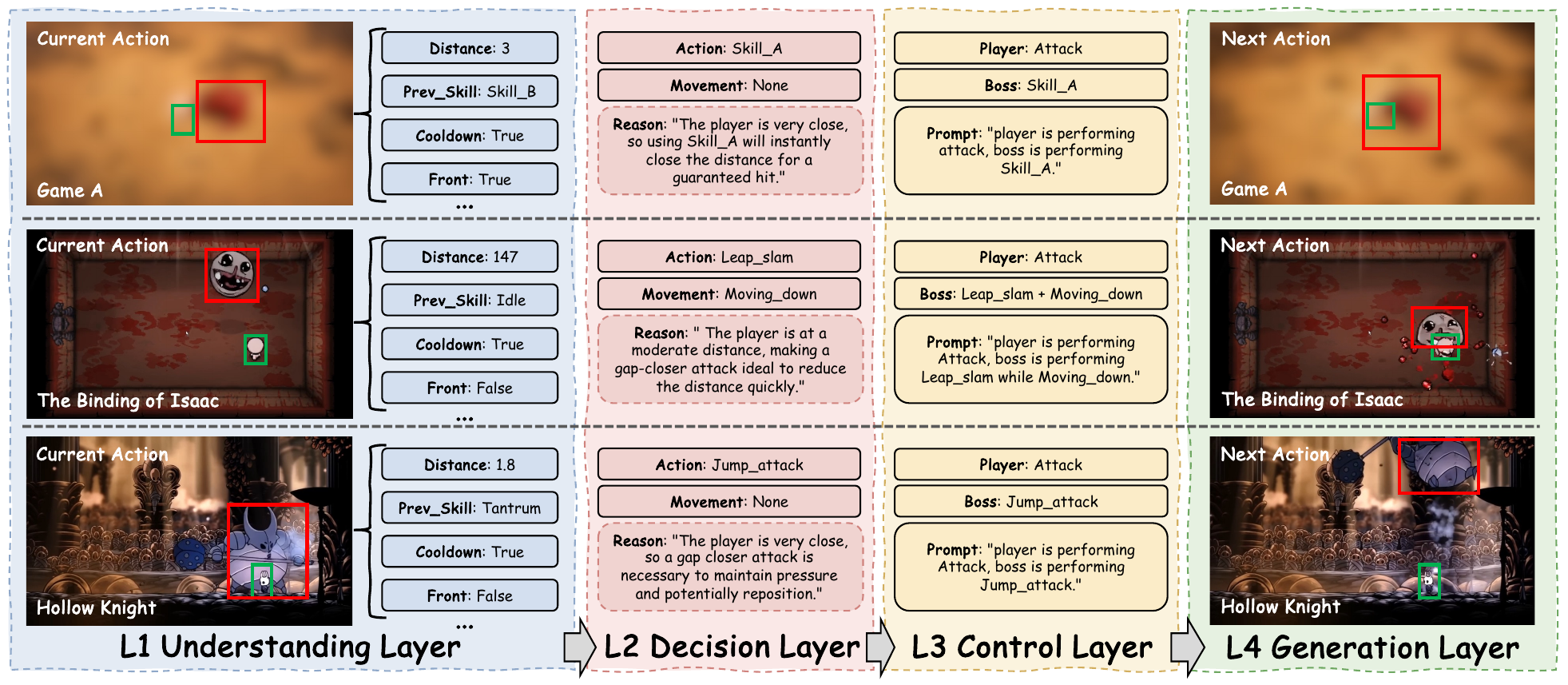}
    \vspace{-1.5em}
    \caption{Qualitative examples of boss behavior grounded in the compact state.
    The top, middle, and bottom rows show \emph{Game A}, \emph{The Binding of
    Isaac}, and \emph{Hollow Knight}, respectively. Each row presents the
    compact state, the resulting boss action and optional
    rationale, the compositional prompt, and the generated outcome.
    Together, the examples show how the framework operates in a closed loop with visual outcomes.
    \textcolor{green!50!black}{Green} and \textcolor{red!75!black}{red}
    bounding boxes denote the player and boss, respectively.} 
    \label{fig:pipeline-qual}
    \vspace{-1em}
\end{figure*}

\section{Experiments}

Our experiments evaluate both the individual layers and the assembled system
by addressing four questions: \textbf{(Q1)} Can L1 accurately
construct the compact state? \textbf{(Q2)} Does L2 ground its reasoning in the compact state and skill mechanics? \textbf{(Q3)} Does the
closed loop produce more responsive, state-aware, and coherent NPC behavior
than implicit or externally controlled baselines? \textbf{(Q4)} Can WorldMind
generalize to games beyond those covered by BOSS-140K?

\subsection{Experimental Setup}
\label{sec:exp-setup}

\noindent\textbf{Games and splits.}
We evaluate L1 and L2 on all three BOSS-140K games; the full-system closed-loop evaluation (Q3) focuses on Game A, whose long-horizon encounters and continuous 2.5D combat provide a demanding test of NPC responses grounded in evolving geometry and skill availability. L1 and L2 use fight-disjoint splits of
frame-aligned decision rows; splits are fight-disjoint. This prevents rows from the same
interaction from appearing in both training and test, avoiding inflated
estimates from near-duplicate decision steps.

\noindent\textbf{L1 configuration.}
L1 uses three causal RGB frames at offsets $[-4,-1,0]$. A pretrained ResNet-18 encodes each frame, and a single-layer GRU
aggregates the sequence. The geometry branch is trained using engine-logged annotations from the dataset and predicts geometry variables from RGB. In parallel, the skill
branch derives skill-history variables deterministically from the boss action
history. Dataset annotations serve only as supervision and only the
geometry branch is evaluated. All learned components are trained once with the seed fixed to 0; reported differences are therefore not accompanied by seed variance.

\noindent\textbf{L2 configuration.}
We instantiate L2 with Gemma-4-E2B-it. At each evaluation point, L1 constructs
the compact state and
skill-history variables. The same skill branch operates on issued actions during closed-loop
inference. Thus, evaluation uses predicted rather than oracle geometry. For the
intervention experiments, L2 receives the cooldown-ready skill menu and
natural-language mechanics descriptions, with skill names replaced by neutral
identifiers while boss identities remain visible. The intervention removes or swaps these descriptions and
evaluates the skill component of the first planned action. The compact state
and cooldown-ready skill menu remain fixed within each pair, isolating the effect of the
mechanics descriptions.

\begin{table}[t]
\centering
\small
\setlength{\tabcolsep}{4pt}
\begin{tabular*}{\columnwidth}{@{\extracolsep{\fill}}lrrrrr@{}}
\toprule
Game & Dist.\ $\downarrow$ & Rel.\ $\downarrow$ & Ang.\ $\downarrow$ & D-bin $\uparrow$ & A-bin $\uparrow$ \\
\midrule
Game A      & 0.312 & 2.9\% & 12.66 & 0.807 & 0.702 \\
Hollow Knight & 0.444 & 6.7\% & 2.96  & 0.772 & 0.996 \\
Isaac         & 8.187 & 8.4\% & 29.38 & 0.904 & 0.839 \\
\bottomrule
\end{tabular*}
\caption{Compact-state reconstruction. Distance is measured on game-specific
scales, while \emph{Rel.} normalizes \emph{Dist.} by each game's median
distance for cross-game comparison.}
\label{tab:l1-fields}
\vspace{-0.5em}
\end{table}
  
\begin{table}[t]
\centering
\small
\setlength{\tabcolsep}{3pt}
\begin{tabular*}{\columnwidth}{@{\extracolsep{\fill}}lrrrrr@{}}
\toprule
Game & $N$ & No-desc $\Delta$ & Swap $\Delta$ & Follow & Stay \\
\midrule
Game A      & 200 & 52.0 & 83.5 & 69.0 & 16.5 \\
Hollow Knight & 200 & 90.0 & 87.0 & 73.5 & 13.0 \\
Isaac         & 121 & 55.4 & 80.2 & 79.3 & 19.8 \\
\bottomrule
\end{tabular*}
\caption{Mechanics-grounded skill selection. \emph{No-desc}/\emph{Swap}
$\Delta$ report first-skill change rates after removing/swapping descriptions.
Under a swap, \emph{Follow} moves to the option carrying the original choice's
mechanics.}
\label{tab:l2-semantic}
\vspace{-1.5em}
\end{table}

\begin{table*}[t]
\centering
\small
\setlength{\tabcolsep}{4pt}
\begin{tabular*}{\textwidth}{@{\extracolsep{\fill}}l l ccc ccc @{}}
\toprule
& & \multicolumn{3}{c}{\textbf{GPT-5.5}} & \multicolumn{3}{c}{\textbf{Gemini-3.1-pro}} \\
\cmidrule(lr){3-5}\cmidrule(lr){6-8}
Method & \shortstack{NPC\\Behavior} & \shortstack{Ours\\Pref.\ (\%) $\uparrow$} & \shortstack{Action\\Valid.\ (\%) $\uparrow$} & \shortstack{Sequence\\Fit (0--5) $\uparrow$} & \shortstack{Ours\\Pref.\ (\%) $\uparrow$} & \shortstack{Action\\Valid.\ (\%) $\uparrow$} & \shortstack{Sequence\\Fit (0--5) $\uparrow$} \\
\midrule
Wan w/o NPC Control & Implicit & 71.5 & 63.6 & 3.28 & 69.8 & 70.6 & 3.20 \\
Wan w/ NPC Control  & Explicit & 70.9 & 63.6 & 3.22 & 70.3 & 70.9 & 3.17 \\
\midrule
\textbf{Ours} & \textbf{State-Aware} & N/A & \textbf{74.0} & \textbf{3.85} & N/A & \textbf{77.6} & \textbf{4.00} \\
\bottomrule
\end{tabular*}
\caption{Closed-loop NPC behavior evaluation. One-minute
\emph{Game A} rollouts evaluated by two LLM judges. All systems use the same
Wan~2.2 backbone and training corpus. Wan w/o NPC Control uses the player-only
generator, whereas Wan w/ NPC Control and WorldMind use the boss-conditioned
generator. All results are averaged over three independent judging passes to
reduce LLM-judge variability. Bold values indicate the best result in each
column.}
\vspace{-0.7em}
\label{tab:main-results}
\end{table*}

\begin{table}[t]
\centering
\small
\setlength{\tabcolsep}{3pt}
\begin{tabular*}{\columnwidth}{@{\extracolsep{\fill}}llrrrr@{}}
\toprule
Game & Method & Dist.\ $\downarrow$ & Ang.\ $\downarrow$ & D-bin $\uparrow$ & A-bin $\uparrow$ \\
\midrule
Game A
 & \textbf{Ours}     & \textbf{0.312} & \textbf{12.66} & \textbf{0.807} & \textbf{0.702} \\
 & DINOv2-S          & 0.722 & 19.16 & 0.601 & 0.606 \\
 & VideoMAE-B        & 0.750 & 16.81 & 0.611 & 0.597 \\
\midrule
Hollow
 & \textbf{Ours}     & \textbf{0.444} & \textbf{2.96}  & \textbf{0.772} & \textbf{0.996} \\
Knight
 & DINOv2-S          & 0.707 & 11.72 & 0.610 & 0.934 \\
 & VideoMAE-B        & 0.830 & 18.60 & 0.549 & 0.836 \\
\midrule
Isaac
 & \textbf{Ours}     & \textbf{8.187} & \textbf{29.38} & \textbf{0.904} & \textbf{0.839} \\
 & DINOv2-S          & 31.17 & 74.04 & 0.642 & 0.613 \\
 & VideoMAE-B        & 38.12 & 84.71 & 0.573 & 0.518 \\
\bottomrule
\end{tabular*}
\caption{Visual encoder ablation. Boss--player geometry 
reconstruction using task-trained and frozen encoders under matched three-frame
inputs.}
\label{tab:l1-main}
\vspace{-1em}
\end{table}

\begin{table}[t]
\centering
\small
\setlength{\tabcolsep}{6pt}
\begin{tabular*}{\columnwidth}{@{\extracolsep{\fill}}lrrr@{}}
\toprule
Game & Text & Text\,+\,Image & $\Delta$ \\
\midrule
Game A      & 69.0 & 69.5 & $+0.5$ \\
Hollow Knight & 73.5 & 67.0 & $-6.5$ \\
Isaac         & 79.3 & 71.1 & $-8.2$ \\
\bottomrule
\end{tabular*}
\vspace{-0.2em}
\caption{Image-modality ablation. Mechanics-following rates (\%) using
text alone or augmented with the frame.}
\label{tab:l2-img}
\vspace{-1.5em}
\end{table}

\noindent\textbf{Full-system evaluation.}
We compare WorldMind with two baselines: \emph{Wan w/o NPC Control}
(\emph{Implicit}) and \emph{Wan w/ NPC Control} (\emph{Explicit}). All three
systems use the same Wan~2.2 backbone and training corpus. Wan w/o NPC Control
uses the player-only generator, which has no boss control channel, whereas Wan
w/ NPC Control and WorldMind use the boss-conditioned generator. The comparison
therefore focuses on how NPC behavior is produced rather than differences in
backbone architecture. To evaluate the appropriateness and tactical coherence
of NPC decisions, we compare the three systems using one-minute rollouts, each
containing 12 decision points. GPT-5.5 and Gemini-3.1-pro evaluate pairwise
preference, per-step action validity and holistic sequence fit. Each judge
evaluates every metric three times, and we report the mean to reduce LLM-judge
variability.

\subsection{L1: Compact-State Reconstruction (Q1)}
\label{sec:exp-l1}

A useful compact state must both faithfully reconstruct the current geometry and retain
information relevant to subsequent decisions. We evaluate these two properties
in turn.

\noindent\textbf{Metrics.}
\emph{Dist.} and \emph{Ang.} are the mean absolute errors (MAEs) of boss--player
distance and relative angle. Distances use engine units, which differ across games, and angles are measured in degrees. \emph{Rel.}
normalizes \emph{Dist.} by the median engagement distance of each game, enabling
comparison across spatial scales. \emph{D-bin} and \emph{A-bin} measure
classification accuracy for fixed distance and angle bins.

\noindent\textbf{Reconstruction fidelity.}
As shown in Table~\ref{tab:l1-fields}, L1 achieves low reconstruction errors across three games. These results show reliable
boss--player geometry reconstruction across viewpoints and spatial scales.
    
\noindent\textbf{Decision relevance of the compact state.}
Accurate reconstruction does not establish whether compacting the state removes
information needed for action selection. We further verify that the compact state retains nearly all
decision-relevant information.

\subsection{L2: Reasoning over the Compact State (Q2)}
\label{sec:exp-l2}

L2 must both ground its choices in the supplied skill mechanics and reason
actions that are tactically appropriate for the current state. We evaluate
these requirements separately.

\noindent\textbf{Grounding in skill mechanics.}
A mechanics-grounded planner should respond when descriptions are removed and
follow the described mechanics when they are reassigned to different skill
identifiers. We test this behavior with a controlled single-skill prompt over
the cooldown-ready menu, isolating skill choice from plan length and ordering.

\noindent\textbf{Metrics.}
All metrics compare the first selected skill with the choice under the full
prompt. \emph{N} is the number of eligible paired decisions.
\emph{No-desc $\Delta$} and \emph{Swap $\Delta$} are the first-skill change
rates after removing or swapping the descriptions. Under a swap, \emph{Follow}
is the rate of selecting the option that now carries the mechanics of the
original choice, whereas \emph{Stay} is the rate of retaining the original
anonymized option.

As shown in Table~\ref{tab:l2-semantic}, removing or swapping
descriptions substantially changes the selected skill. Under swaps, the model
follows the transferred mechanics more often than it retains the original
identifier, showing that its decisions are grounded in skill mechanics rather
than fixed names.

\noindent\textbf{Tactical quality.}
Mechanics grounding does not by itself establish tactical appropriateness.
We further evaluate that L2's planned skills are consistently rated as more
appropriate than random legal choices from the same cooldown-ready menu,
showing that the compact state informs tactical selection beyond action
legality.

\subsection{Full Closed-Loop System (Q3)}
\label{sec:exp-gen}

\noindent\textbf{Metrics.}
\emph{Ours Pref.} is the fraction of pairwise comparisons in which a judge
favors WorldMind over a baseline. \emph{Action Valid.} is the percentage of
individual decisions judged valid for the current situation, while
\emph{Sequence Fit} rates the tactical coherence of the complete rollout on a
$0$--$5$ scale.

As shown in Table~\ref{tab:main-results}, WorldMind is preferred over both
baselines in approximately $70\%$ of pairwise comparisons and ranks first in
action validity and sequence fit under both judges. This consistent advantage
shows that the closed loop system improves NPC behavior at both the action and
long-horizon sequence levels.

\subsection{Cross-Game Generalization (Q4)}
\label{sec:exp-generalization}

On the publicly released WildWorld dataset~\cite{wildworld}, the Decision Layer shows partial cross-game generalization and remains state-sensitive, whereas compact-state reconstruction requires target-domain adaptation.

\subsection{Ablation Studies}
\label{sec:exp-ablation}

We conduct ablations across the framework and highlight its two most important
design choices here: the visual encoder for L1 and the input modality for L2.

As shown in Table~\ref{tab:l1-main}, the task-trained L1 encoder outperforms
the frozen DINOv2-S~\cite{dinov2} and VideoMAE-B~\cite{videomae} encoders on
every reconstruction metric. In Table~\ref{tab:l2-img}, adding the rendered frame to the
compact-state text provides no consistent improvement. This also provides indirect evidence that the compact state
is sufficient for L2's mechanics-grounded decisions.

\section{Conclusion}

We introduced WorldMind, the first decoupled game world model that makes NPC behavior an
explicit part of interactive world modeling. WorldMind separates compact-state
reconstruction, NPC reasoning and planning, action control, and video
generation, then reconnects them through a closed loop in which each generated
outcome informs the next NPC action. Experiments demonstrate our NPC behavior responds
coherently to the evolving game state, rather than remaining implicit in video
generation or being prescribed externally.

\section{Acknowledgments}

We would like to express our sincere gratitude to Ruidong Wang and Murphy Zhao for their tremendous support throughout this project.

\bibliography{references}

\end{document}